# FLOW OF ACTIVITY IN THE OUROBOROS MODEL

Knud Thomsen
*Paul Scherrer Institute*
*CH-5232 Villigen PSI*
email: knud.thomsen@psi.ch

**Abstract**

*The Ouroboros Model is a new conceptual proposal for an algorithmic structure for efficient data processing in living beings as well as for artificial agents. Its central feature is a general repetitive loop where one iteration cycle sets the stage for the next. Sensory input activates data structures (schemata) with similar constituents encountered before, thus expectations are kindled. This corresponds to the highlighting of empty slots in the selected schema, and these expectations are compared with the actually encountered input. Depending on the outcome of this "consumption analysis" different next steps like search for further data or a reset, i.e. a new attempt employing another schema, are triggered. Monitoring of the whole process, and in particular of the flow of activation directed by the consumption analysis, yields valuable feedback for the optimum allocation of attention and resources including the selective establishment of useful new memory entries.*

## 1. INTRODUCTION

The Ouroboros Model proposes a general algorithmic layout for efficient self-steered data processing in agents [1]. A very coarse characterization of the involved processes along with selected consequences for living beings has been given before [2]. It can be claimed that the Ouroboros Model fares well in the light of venerable criteria posited for artificial intelligence and that it serves nicely as a starting point for explaining the emergence of consciousness [3]. In the following, a somewhat more technical description of the Ouroboros Model and the flow of activity from one step in the data processing to the next are presented.

This is work in progress; it aims at understanding general intelligence, and even consciousness in the end, starting from a new perspective and following a top-down engineering approach. In this short note a conceptual design is outlined in four steps devising the algorithmic structure of the Ouroboros Model. The focus lies on the flow of data processing in rather broad terms and not on exact formalization and neither on low level feature extraction, on grounding, embodiment or physical action. In a sense, the here sketched description now only afterwards provides the basis for the claimed effects outlined earlier [2,3,4,5].

A truly fundamental requirement for consistency, most notably consistency between experience, action and perception, lies at the heart of the Ouroboros Model [2]. As a direct consequence, the proposed algorithm funnels all unfolding activity of an agent repeatedly through one stage where overall consistency of all current (neural) activation is checked. The end of one processing cycle is at the same time the beginning of the next iteration while its results provide the new basis: the snake of data-processing devours its tail.

## 2. DESIGN IN FOUR STEPS

### 2.1. Basic Loop

A principal activity cycle is identified; starting with a simple example of sensory perception the following succession of steps can be outlined:

> ... anticipation
> perception
> evaluation
> anticipation …

The identified data processing steps, i.e. sub-processes, are itemized and briefly discussed below. They are linked into a full circle as shown in Figure 1.

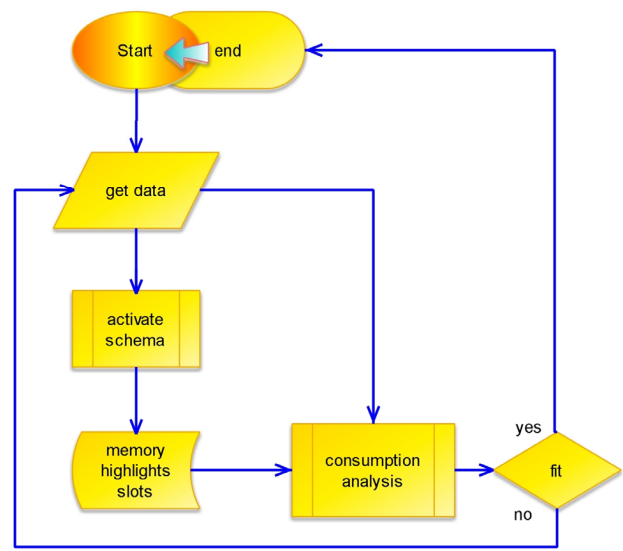

Figure 1. Basic loop structure of the Ouroboros Model.

**Start:**
This is the almost arbitrary entry point in the perpetual flow of the proposed data-collection and -evaluation processes: a novel episode commences with little heritage from previous activity.

**Get data:**
In this example first perceptional data arrive as input.

**Activate Schema:**
Schemata are searched in parallel; the one with the strongest bottom up activation sharing similar features is activated.

**Memory highlights Slots:**
Each of the features making up the selected schema are marked as relevant and they are activated to some extent; this biases all features belonging to this schema also when they are not part of the current input, i.e. empty slots are thus pointed out.

**Consumption Analysis:**
This is the distinguished recurrent point at the core of the main cyclic process constituting the Ouroboros Model.

A comparison of the demanded attributes of the activated schema with the actually available features will often lead to satisfactory correspondence; the current cycle thus is concluded without gaps, and a new processing round can start.

If the achieved fit is not sufficient, e.g. slots are left unfilled, follow-up action is triggered. In the outlined most simple example, more data are searched for, guided by expectations in the form of the biased empty slots.

**End / new Start:**
In the example of Figure 1. a (preliminary) end is reached when good agreement between expectations and data is detected, e.g. an object is recognized; a new episode can start.

The current emphasis on data processing leads to the neglect of any other, in particular bodily, action, - even if movements often are of highest importance to living creatures. Obviously, the information that expectations based on experience are in accord with current sensory data is useful for any living being as well as artificial agent, no need for action is signaled then. In case some discrepancy is detected, it might often be wise to collect further data as a first step.

Recently, in a state of the art model of image interpretation the substantial advantage of combining a bottom-up and a top-down pass into a cycle has been demonstrated [6].

## 2.2. Extended Loop

The basic loop of Figure 1. does not offer much room for sophisticated data processing or possibilities for growth. At the very minimum provisions for applying different memories and a learning mechanism to establish new ones has to be included.

If the first selected schema does not lead to any satisfactory fit, another script, one that was disregarded first, has to be tried. In case no preexisting schema can accommodate the sensory data well enough, a new data structure is generated, such, that at least during the next encounter of a similar situation, relevant memories can be brought to bear.

The basic loop can easily be extended with flexible schema selection and memory capabilities as indicated in Figure 2. The most notable addition is a "reset" process. Consumption analysis in most cases will not deliver a clear cut yes / no decision. The range of achieved correspondence can vary from very bad to perfect.

**Reset:**
If nothing fits ("impasse"), a "reset" is triggered and the cycle starts anew, this time with another schema and avoiding the first one. The assignment ("consumption") of the available sensory input data is investigated with respect to a second schema while the originally selected one is bypassed and muted. Reciprocal inhibitory links between schemata in the Figure 2. indicate a winner take all competition between possibly applicable schemata; at the same level only one can be active at a given time.

Assuming for the time being a static scene as the source of input, never-ending minor discrepancies also will eventually cause a reset; - another schema will be tried after a number of unsuccessful iterations. In the last chapter it will be suggested how this threshold can itself be determined in an adaptive way as a result of feedback in the context of relevant experience.

Even if no mismatch is detected, when repeating the loop too often without much new input a timeout will cause the switching to another schema; for ambiguous figures the perception will flip after some time [5].

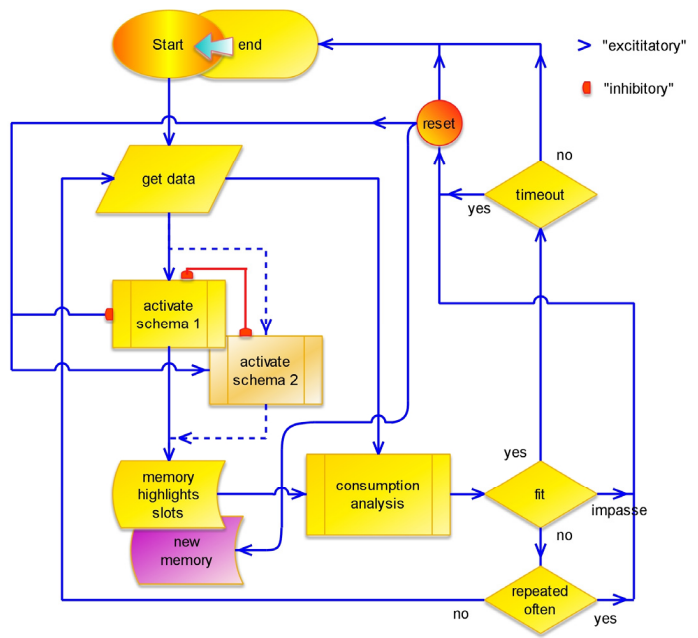

Figure 2. Basic loop augmented with mechanisms for flexible schema selection and the recording of likely useful new memories. Although connections are marked "excitatory" and "inhibitory", no direct correspondence to nervous structures is intended at the moment; "excitatory" simply stands for a link activating the receiving entity, and "inhibitory" means that arriving activation dampens or prohibits activity of the terminal process.

**New Memory:**
Whenever a reset is triggered this marks an important point in the flow of activity. If the interruption arose from an impasse it is of advantage to record the status of affairs for future avoidance of that dead end. A new memory entry, i.e. a novel schema connecting the features occurring just before the redirection of the process can provide the basis for better assessment when encountering similar input the next time: features can be assigned faster and action can be triggered already after less iterations.

Also at the occasion that everything fits nicely it certainly is useful to establish an entry into memory which connects the antecedent activity and the successful outcome into a new schema, which will facilitate the same achievement later again. As soon as some of these features are active afresh, the activation of that schema will bias the remainder and thus highlight all steps leading to a repeated success.

In any case it is important to note that the activations, which are knit together into a new schema, will include detailed representations of the last steps that led to a distinct result. At later occurrences in similar circumstances these memories will serve for defining goals directing promising activity.

## 2.3. Self-directing Loop

In addition to the immediate feedback on the quality of fit between expectations and actual input provided by the consumption analysis directly, (meta-) information can be obtained by monitoring the general flow of action around the circuit as indicated in Figure 3.

Time is an important parameter, any input will become available in a piecemeal manner, even for static stimuli, as different senses do not all work at the same pace. Incrementation of data over time is the rule even more for changing inputs e.g. ones including movements.

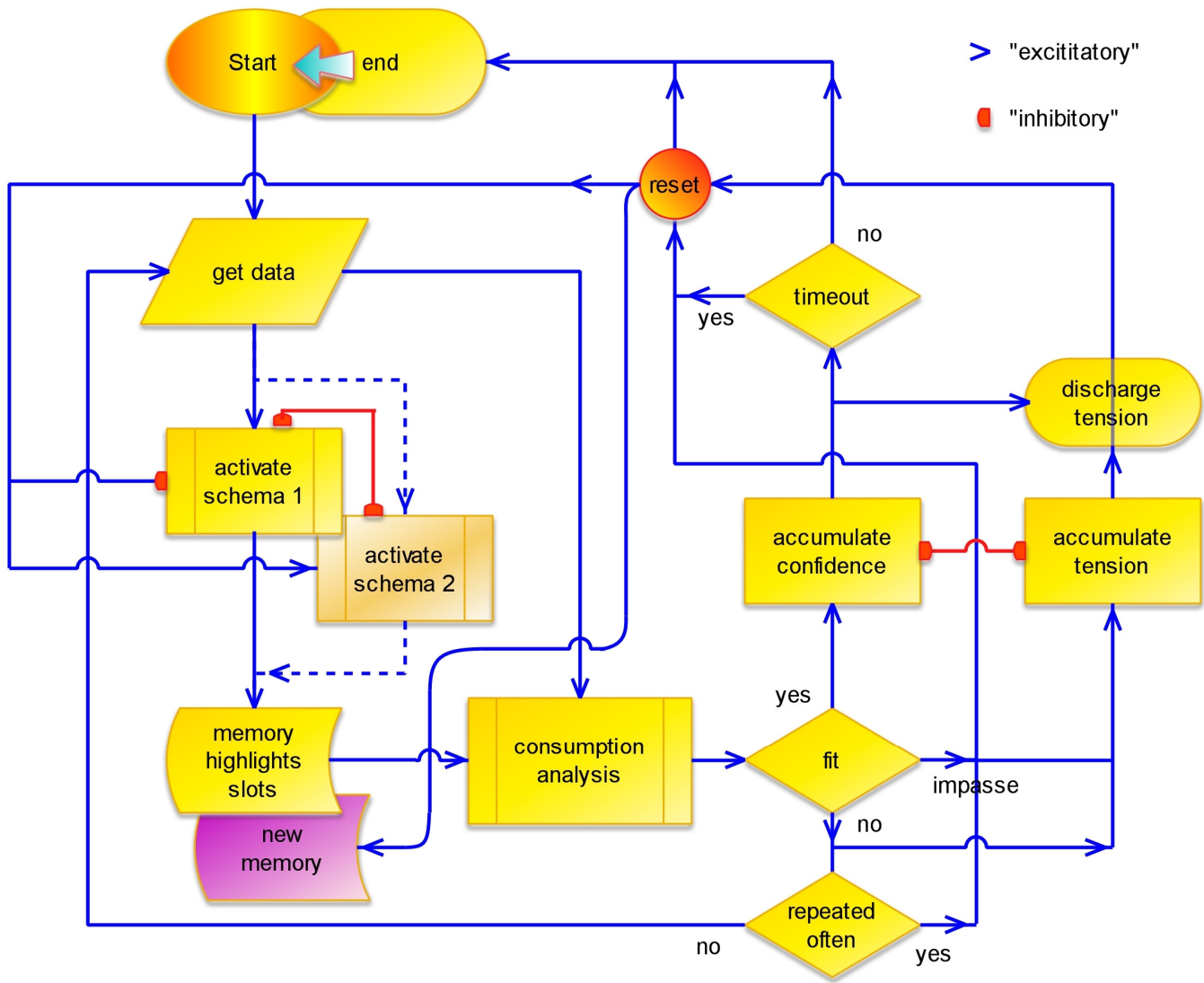

Figure 3. The process flow in the loop allows for self monitoring by keeping track of the current performance.

Steady progress, i.e. an increase in the quality of the overall fit from one iteration of the loop to the next, definitively is a good result. Modules for keeping track of the development can be built based on the integration of interim outcomes of the consumption analysis.

**Accumulate Confidence / Tension:**
When a scene develops over time also under the very best conditions its perception and interpretation will not be finished in one moment after one turn around the loop. With each acceptable fit of a part of the input, confidence in the selected schema, the anticipated script, will increase, and only minimal tension will build up.

Conversely, a changing target makes it difficult to achieve a final interpretation. Repeated trials without success will increase the pressure on the agent to find a satisfactory solution and tension accumulates.

When eventually a fitting structure is found after several frustrated attempts, the piled-up tension will be released. The ensuing reset will be the more pronounced the more tension had built up.

An impasse or high accumulated frustration can trigger a reset without relieving all the stress.

This dynamic flow of activity is not limited to genuinely dynamic scripts featuring some movements or other changes over time. Generally, with limited input rates, also static scenes or objects will actually be parsed in steps over an extended period of time. Experiments with observers who were viewing pictures have repeatedly demonstrated canonic scan paths of the eyes leading to the complete interpretation of the picture. Saccades are made in sequences jumping from one informative point to the next [7]. For complex images top down guidance of the data foraging by the gist of the scene has been suggested [8].

The Ouroboros Model naturally explains these findings because a schema, e.g. for a face, will be activated by the first available data, e.g. a feature looking like an eye, and then the biased slots for other expected attributes, like an ear or a nose will draw the attention of the observer to the most likely locations where they are expected. These will be found at the positions there is some activation building up from the beginning. Bottom-up and top-down activities thus cooperate to achieve an overall consistent interpretation efficiently and quickly. Adding slots for temporally changing dimensions in the input data allows taking care of movements and other developments in just the same way.

Consumption analysis not only directs the process of interpretation / analysis and the allocation of attention during the ongoing data processing, it also delivers the signal when to stop because either a scene has been understood satisfactorily and resources can be devoted to other issues or because further processing with the available information is not considered promising.

The computational power of the Ouroboros Model derives from the interplay between data structures and the processing steps based on them. Its potential unfolds when the activity and its results at one point in time lead to improved structures for future use; i.e. new memories are laid down, new associations and schemata are learned. Accumulated confidence and tension serve effectively as control signals modulating the significance of a reset and the importance of remembering a specific episode.

Memories are laid down gradually when inconspicuous sequences are repeated often or, instantaneously at only one occasion, when much tension is involved. This measure of importance will be included in the new schema.

As already Otto Selz, who introduced the schema-concept a century ago, remarked, any schema or part of it can serve as the building block for new schemata [9]. These structures thus are nested into hierarchies. They are not restricted to static snapshots but they can contain dynamic components, i.e. representations of transients.

There are slots in schemata for features in sensory data and also for signals from the agent herself, and they can include information on the normal course of progresses that the actor usually makes in similar data processing [4].

On top of monitoring ongoing perceptual performance, consumption analysis then also allows for the assessment of the status and the achievements very tightly linked to the agent with details relating to his embodiment.

What has been described above taking the example of perception, basically applies equally for other actions like active movements of the agent.

### 2.4. Full Loop in the Ouroboros Model

The last step toward the "full loop" as envisaged by the Ouroboros Model can only be sketched, even more so than the other three stepping stones presented above; "full" as presented here, definitively does not mean that no more extensions would be possible in the proposed framework.

In a comprehensive formulation of the Ouroboros Model grounding and embodiment are essential for explaining details of the implementation and the performance of living creatures [2]. First feature detectors like for faces and associations as an inborn fear of snakes have been selected as useful for survival over the course of evolution. Even without directly considering this basis the full loop model can shed some light on the self-reflective nature of high level data processing. Self-monitoring is identified as a foundation and origin of general intelligence, irrespective whether natural or artificial.

The power and flexibility of the Ouroboros Model is based on the interlacing of the different processes around consumption analysis and the derived signals. By establishing self-reflective schemata including feedback and status-signals, (meta-)information can be employed for the adaptive steering of all activity. Meta-monitoring and control is possible by applying the same steps and processes as for simple perception, - only with more sophisticated schemata and concepts. Overall control is exercised on a different time scale than the one of single activation loops.

Tension and confidence signals are built up over several iterations and they will also be effective over longer periods. These signals are some type of summary outcome of all short-lived activity and they bias every subsequent action at the same time. It has been argued that in living such phenomena they are known as emotions [2].

Some prominent interactions between stages are indicated in Figure 4. The connections are in all cases reciprocal and they act more as modulators influencing other signals than triggering directly actions on their own.

As one example, the time, i.e. number of trials, it takes before a reset is triggered even in case of an acceptable fit ("time-out") will be determined by the applicable schema, the context, how many other possibly fitting schemata are available, and by the general arousal, corresponding to the level of tension. The other way round, frequent interrupts will raise the excitation and tension. To start with, schemata, which contain commensurate values for the "emotional" dimensions will preferentially be chosen first.

The schemata in memory on their part determine the expected magnitude of associated arousal.

In living actors the above status and performance signals are grounded and tied very tightly to their body.

## 3. RELATION TO OTHER WORK

The Ouroboros Model can probably be best conceived as a continuation of the work by Otto Selz bending his linearly ordered chain of steps to a full ring making the snake bite its tail [9]. Obviously, the next steps for the Ouroboros Model have to investigate its relation to extant models, in particular, to conceptions developed in 50 years of research in Artificial Intelligence on one side and to models of brain function based on neurological mechanisms on the other.

### 3.1. Artificial Intelligence

Schemata, frames or scripts are nothing new and data structures of such type are employed by many approaches, production systems being the most prominent ones [10]. The main innovation of the Ouroboros Model is that it offers a self organizing *recursive* account, where truly self-steered learning progresses in iterative steps.

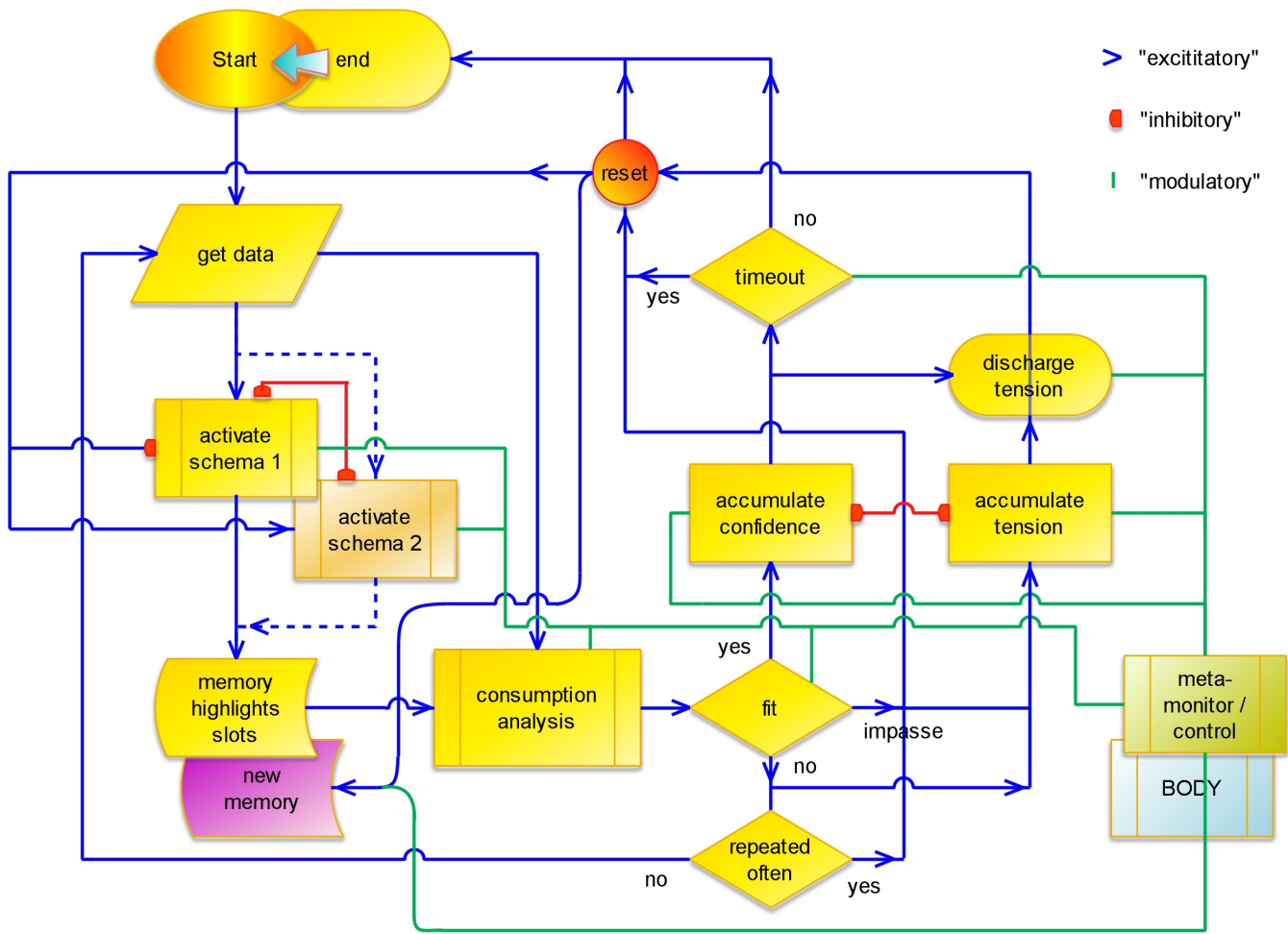

Figure 4. In the full Ouroboros Model self-monitoring is employed for self-steering; the flow of activity and the outcome of the different processing stages exert mutual modulations.

In the Ouroboros Model results from any one data processing step have important impact on all subsequent activity. Recursion is providing computational power and self-reference means more than just lip service. Starting from some predefined structures, new, and particularly useful ones are generated incrementally, connecting at points where the need surfaces. Thus self-expanding cognitive structures are continually refined and further expanding.

It can be argued that the interplay between the schemata available at one point in time and the then newly arriving input forms a somewhat peculiar implementation of optimum Bayesian combination of all relevant information.

The Ouroboros Model at the same time can be seen as a direct extension of production systems, with linear if→then rules turning out to be particular instances involving schemata with a single slot left open.

### 3.2. Natural Brains

Given that, despite all efforts of AI, living brains in total are still the most effective and versatile data processors that we know of, it should be shown how structures and processes described by the Ouroboros Model map to entities found in animal and human brains.

At this point only two conjectures shall be made: the hippocampus, and in particular the dentate gyrus, appear to be well suited for taking snapshots of all brain activity at specially marked occasions thereby generating index-entries which link specific features spread over wide areas into unique concepts, i.e. schemata. The anterior cingulate cortex would then be the region where discrepancies between expectations and actual findings are determined as described by consumption analysis.

In the end, the Ouroboros Model should account for neurological findings and also shed light on the highest levels of cognition and consciousness [11]; if the here presented suggestions are useful and reflect some reality, it should be possible to explain the so far established models as special cases when emphasizing some specific selected dimensions.

As one major claim of the Ouroboros Model it shall only be mentioned that from the very essence of the suggested algorithmic structure and it's working under almost constant time pressure (striving for survival in a real world) it follows that inevitably "left-overs" are produced all the time, i.e. features unaccounted for as well as half-filled schemata. It has been proposed that this is the very reason why all brains exhibiting non-trivial intelligence appear to depend on sleep [1]. The main function of sleep and dream according to the Ouroboros Model is to maintain a healthy signal to noise ratio, partly by actively disposing of data garbage accumulating during normal awake activity.

## 4. CONCLUSIONS

The Ouroboros Model offers a novel self-consistent and self-contained account of efficient self-reflective data processing in self-steered agents.

A principal problem for exposing the somewhat involved concepts and their interdependencies lies in the principal cyclic nature of the depicted processes. Seemingly circular connections and arguments are easily avoided by carefully observing the direction of time: any processing step relies only on data and structures available from before but it works "backwards" in the sense that it has the power to influence these items leading to changes, which then become effective for the same process but only afterwards during subsequent process cycles.

It goes without saying that a lot of work is still needed to substantiate the here presented proposals and conjectures and to establish the Ouroboros Model as the universal algorithmic backbone for efficient cognition. If this paper sets a frame and points towards promising directions of future research it serves its purpose and this hopefully in a comprehensively self-consistent manner.